\documentclass[11pt]{article}

\usepackage[preprint]{acl}

\usepackage{times}
\usepackage{latexsym}
\usepackage{amsmath}
\usepackage{multirow}
\usepackage{makecell}
\usepackage{tcolorbox}
\usepackage[T1]{fontenc}

\usepackage[utf8]{inputenc}
\usepackage{amssymb}
\usepackage{booktabs}
\usepackage{multirow}
\usepackage{subcaption}
\usepackage{algorithm}
\usepackage{algorithmic}

\usepackage{microtype}

\usepackage{inconsolata}

\usepackage{graphicx}

\title{%
\makebox[0pt][r]{%
  \raisebox{-0.45\height}[0pt][0pt]{%
    \includegraphics[width=1.3cm]{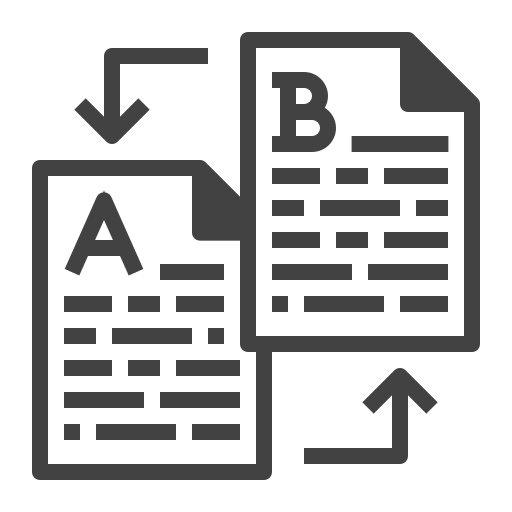}%
  }%
  \hspace{1mm}
}%
Refinement Provenance Inference: Detecting LLM-Refined \\
Training Prompts from Model Behavior}

\author{Bo Yin\thanks{Equal contribution.}$\quad{}$ Qi Li\footnotemark[1]$\quad{}$ Runpeng Yu$\quad{}$ Xinchao Wang\thanks{Corresponding author.} \\
National University of Singapore  \\
\texttt{\{yin.bo, liqi\}@u.nus.edu$\quad{}$ xinchao@nus.edu.sg }}

\begin{document}
\maketitle

\begin{abstract}

Instruction tuning increasingly relies on LLM-based prompt refinement, where prompts in the training corpus are selectively rewritten by an external refiner to improve clarity and instruction alignment. This motivates an instance-level audit problem: for a fine-tuned model and a training prompt–response pair, can we infer whether the model was trained on the original prompt or its LLM-refined version within a mixed corpus? This matters for dataset governance and dispute resolution when training data are contested. However, it is non-trivial in practice: refined and raw instances are interleaved in the training corpus with unknown, source-dependent mixture ratios, making it harder to develop provenance methods that generalize across models and training setups.
In this paper, we formalize this audit task as Refinement Provenance Inference (RPI) and show that prompt refinement yields stable, detectable shifts in teacher-forced token distributions, even when semantic differences are not obvious. Building on this phenomenon, we propose RePro, a logit-based provenance framework that fuses teacher-forced likelihood features with logit-ranking signals. During training, RePro learns a transferable representation via shadow fine-tuning, and uses a lightweight linear head to infer provenance on unseen victims without training-data access. Empirically, RePro consistently attains strong performance and transfers well across refiners, suggesting that it exploits refiner-agnostic distribution shifts rather than rewrite-style artifacts. 
The code is available at: \url{https://github.com/YinBo0927/RePro}.

\end{abstract}

\section{Introduction}
\begin{figure}[t]
    \centering
    \includegraphics[width=\linewidth]{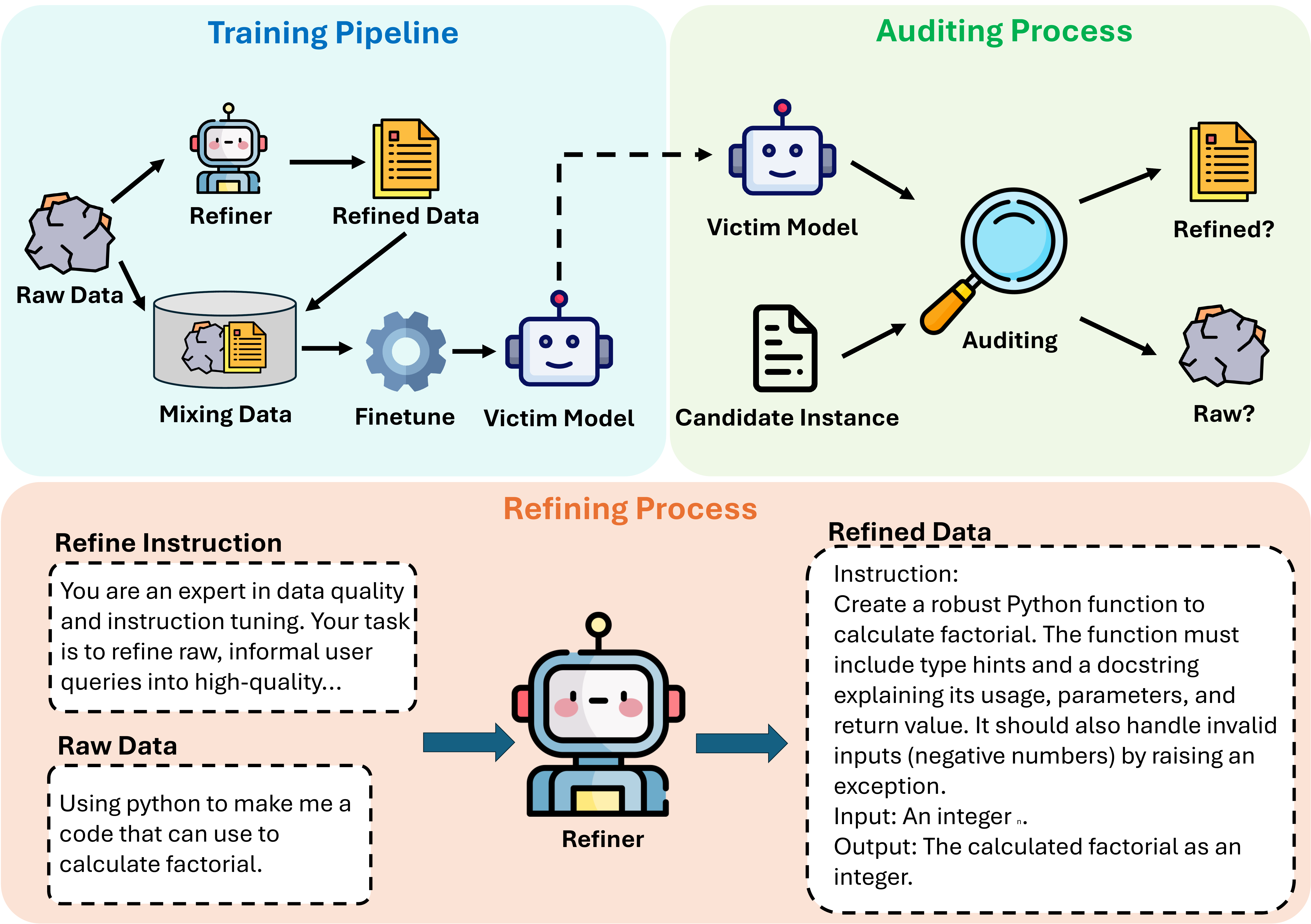}
    \vspace{-6mm}
    \caption{The process of Refinement Provenance Inference (RPI) problem.}
    \label{fig:example}
    \vspace{-6mm}
\end{figure}
Large language models have rapidly evolved into general-purpose systems that power a wide range of applications, from reasoning and code generation to dialogue and tool use~\cite{achiam2023gpt,dubey2024llama,team2023gemini,roziere2023code}. As capabilities have improved, fine-tuning and instruction tuning have become standard practices for adapting these models to specific domains and interaction styles, which in turn has placed increasing emphasis on the construction and curation of high-quality training prompts~\cite{yin2025fera, yin2025don,ouyang2022training, hu2022lora, li2023towards}. In many modern pipelines, raw prompts, the original collected instructions, are rewritten by a refiner model to standardize phrasing, reduce ambiguity, and align with instruction-following conventions~\cite{xu2024wizardlm,mukherjee2023orca,yan2025lacadmlatentcausaldiffusion, liu2024improving}. This widely used refinement step raises a provenance question for auditing: given a fine-tuned model and a candidate instance $(x_j, y_j)$, \textbf{can we infer whether the model was tuned on its raw version or on its refiner-rewritten counterpart?}

Answering this question matters for both transparency and risk assessment in model development~\cite{longpre2023data,mitchell2019model}. From an auditing perspective, refinement can materially change the distribution of training prompts, and practitioners may wish to verify whether a deployed model was trained under a declared data pipeline or whether an undisclosed refiner was used~\cite{mokander2024auditing,dziedzic2022dataset}. From a security and privacy perspective, the refinement step may also act as a distinctive transformation that leaks information about the training process itself, potentially revealing aspects of an organization's data preparation workflow~\cite{carlini2022quantifying,nasr2023scalable, li2025towards}. 

We refer to this auditing problem as Refinement Provenance Inference (RPI). Figure~\ref{fig:example} shows the process of RPI. Notably, this is a data-level provenance problem: the victim may have been fine-tuned on a mixture of refined and raw prompts, and the goal is to localize which training instances were refined. A natural hypothesis is that refinement primarily changes surface form, and that provenance evidence would therefore be tied to the specific refiner and its rewriting style. However, we argue that \textbf{training on refined prompts induces distribution-level preference shifts that persist beyond surface realizations.} Concretely, refinement tends to make prompts more canonical and better aligned with instruction-following conventions, which biases the gradients observed during fine-tuning and can alter the victim model's token-level preferences under teacher forcing~\cite{shumailov2023curse,zhou2023lima,santurkar2023whose,gudibande2023false}. These shifts are not always obvious from generated text, but they can be measured from the teacher-forced token distributions as changes in likelihood patterns, ranking behavior among top candidates, and logit margins~\cite{shi2023detecting,hans2024spotting,gonen2023demystifying}. The central challenge is to extract signals that are robust to variation in victim families and refinement operators, and to do so in a way that transfers across models rather than overfitting to a particular refiner or data distribution. 

To address this challenge, we propose \textbf{RePro}, a logit-based framework for refinement provenance inference that learns transferable signals in a shadow training setup. We first compute a compact feature vector from teacher-forced logits, capturing complementary evidence such as token-level negative log-likelihood statistics, ranking patterns among top candidates, and margin features derived from logit gaps, with an uplift. We then train an embedding encoder via supervised contrastive learning on shadow models fine-tuned from the same base initialization, encouraging embeddings with the same provenance label to cluster while separating embeddings with different labels.
Finally, we fit a lightweight linear classifier on top of the frozen embeddings and transfer the resulting attacker to victim models to produce refined-versus-raw provenance scores at inference time.

Our contributions are as follows:
\begin{itemize}
  \item \textbf{New provenance task.} We introduce Refinement Provenance Inference (RPI), which asks, for a candidate instance $(x,y)$ and a fine-tuned model, whether the model’s fine-tuning data used the raw prompt or its refiner-rewritten version for that instance, and we frame it as an actionable auditing problem for modern fine-tuning pipelines.
  \vspace{-2mm}
  \item \textbf{Provenance framework.} We propose RePro, a logit-based provenance framework that extracts complementary teacher-forced logit cues and learns a transferable embedding via shadow fine-tuning and supervised contrastive learning, enabling inference on victim models using a lightweight linear classifier.
\vspace{-2mm}
  \item \textbf{Evidence for detectable and transferable traces.} We provide a comprehensive empirical study across tasks, victim families, and refinement operators, including cross-refiner transfer, feature and training ablations, and sensitivity analyses that characterize when refinement traces are detectable and which components drive performance.
\end{itemize}

\section{Related Work}
\label{sec:related}

\subsection{Training-Data Auditing}
A long line of work shows that training induces systematic changes in a model's confidence landscape that can be exploited for auditing~\cite{shokri2017membership,yeom2018privacy,song2021systematic,salem2018ml}. In membership inference, attackers distinguish seen versus unseen examples using statistics such as loss, entropy, or margin, and stronger variants rely on shadow-model transfer, calibration features, or query-efficient probing~\cite{carlini2022membership,duan2024membership,ko2023practical, li2025vidsmemembershipinferenceattacks}. Beyond membership, property inference predicts whether the training set contains examples with a particular attribute by aggregating output statistics, highlighting that model outputs can leak training-time signals even when the attribute is not directly observable from the generated text~\cite{ateniese2015hacking,kandpal2024user,ganju2018property,mahloujifar2022property}. For language models, studies on memorization and data extraction further support that token-level likelihood patterns can encode training-time regularities~\cite{carlini2022quantifying,shi2023detecting}. Our work follows this general paradigm but targets a different training attribute, namely whether a fine-tuning instance was presented in an LLM-refined form rather than its raw form, which motivates logit-centric signals and transfer-based attackers instead of text-only evidence.
\vspace{-1.5mm}
\subsection{Data Refinement and Detection}
LLM-driven rewriting is now widely used in large-scale data curation, particularly for instruction-tuning where prompts are standardized, clarified, and aligned to target interaction styles~\cite{xu2024magpie,ding2023enhancing,li2024datalineageinferenceuncovering}. Prior work studies refinement operators and policies, showing that automated rewriting can shift both surface form and latent preferences, yielding refined distributions that systematically differ from raw data~\cite{lee2023rlaif,sun2023principle, li2023hept}. While refinement is typically treated as a quality-improving preprocessing step, its downstream footprint as an auditable training attribute has received less attention~\cite{golchin2023time,zhang2024fine,lyu2024discussion}. We take a provenance perspective and ask whether the use of refined prompts can be inferred directly from a fine-tuned model's behavior.

A related line of research aims to detect machine-generated or machine-transformed text via likelihood artifacts, perturb-and-score stability tests, stylometric signals, and watermarking~\cite{mitchell2023detectgpt,yang2023dna,su2023detectllm,kirchenbauer2023watermark,kuditipudi2023robust}. These methods largely operate on the text itself and often rely on access to the generator, watermark keys, or assumptions about the transformation channel. Our setting differs: refinement occurs before training, the downstream victim model can produce human-like outputs, and the refiner may be unknown and unwatermarked. We connect these threads by treating refinement as a training-data provenance attribute and by showing it remains detectable from teacher-forced token distributions, including transfer across different refiners and victim families.

\section{Refinement Provenance Inference}
\label{sec:problem}
\subsection{Problem Definition}
Modern fine-tuning pipelines often refine training prompts using an external LLM to improve clarity and consistency.
Such refinement can induce systematic shifts in the effective training distribution, which may be reflected in the token-level predictive behavior of the fine-tuned model.
We study refinement provenance inference at the instance level: within a single fine-tuned victim, different training instances may use different prompt variants, and the goal is to infer, for each instance, whether the prompt used during fine-tuning was raw or LLM-refined.
We emphasize that refinement is applied only to the input prompt, while the reference output remains unchanged.

We index semantic instances by $i$, where each instance corresponds to a unique underlying task with a raw prompt $x_i^{\mathrm{raw}}$ and a reference output $y_i$.
A refinement operator $R(\cdot)$ maps the raw prompt to a refined prompt:
\begin{equation}
x_i^{\mathrm{ref}} = R\!\left(x_i^{\mathrm{raw}}\right).
\end{equation}
A fine-tuning dataset is constructed by selecting, for each instance $i$, either the raw or refined prompt with an i.i.d.\ latent indicator $z_i \sim \mathrm{Bernoulli}(\rho)$:
\begin{equation}
x_i^{\mathrm{tr}} = x_i^{z_i} =
\begin{cases}
x_i^{\mathrm{ref}}, & z_i=1,\\
x_i^{\mathrm{raw}}, & z_i=0,
\end{cases}
\qquad z_i \in \{0,1\},
\label{eq:mixture}
\end{equation}
where $z_i$ is the refinement provenance label ($1$ for refined, $0$ for raw).
Let $M_0$ denote a base language model and $M_a$ denote the victim obtained by supervised fine-tuning (SFT) on the mixture:
\begin{equation}
M_a \leftarrow \mathrm{SFT}\!\left(M_0;\ \{(x_i^{\mathrm{tr}}, y_i)\}_{i \in \mathcal{I}_a}\right),
\label{eq:victim}
\end{equation}
where $\mathcal{I}_a$ is the set of semantic instances used to fine-tune the victim.

\noindent
\textbf{Auditing task.}
For an instance $i \in \mathcal{I}_a$ (membership known), the auditor is given the victim $M_a$ and an evaluation pair $(\tilde x_i, y_i)$ for teacher forcing, and aims to infer the training-time provenance label $z_i$:
\begin{equation}
s_i = g\!\left(\phi(M_a; \tilde x_i, y_i)\right) \in [0,1], \qquad \hat z_i = \mathbf{I}[s_i \ge \tau],
\label{eq:predict}
\end{equation}
where $\phi(\cdot)$ extracts features from the victim's token-level predictive behavior on $y_i$ conditioned on $\tilde x_i$, $g(\cdot)$ outputs a classification score, and $\tau$ is a threshold.
In our main setting, $\phi(\cdot)$ is computed from teacher-forced log-probabilities and top-$k$ logit statistics.

\subsection{Access Assumptions}
\label{sec:access}
We assume an auditor has (i) query access to a fine-tuned victim model $M_a$; (ii) an evaluation set of instances with reference outputs; and (iii) access to the underlying base model $M_0$ to construct shadow fine-tuned models for learning transferable decision rules. Given a candidate pair $(x_j,y_j)$, the auditor performs teacher forcing on $y_j$ to obtain token-level log probabilities and compute NLL-based statistics. Our main setting further assumes access to top-$k$ logits. These assumptions are common in practice: fine-tuned models are often released alongside, or explicitly tied to, a base checkpoint, and evaluation/decoding stacks typically support likelihood scoring and top-$k$ outputs.

\begin{figure}[t]
  \centering
  \includegraphics[width=\linewidth]{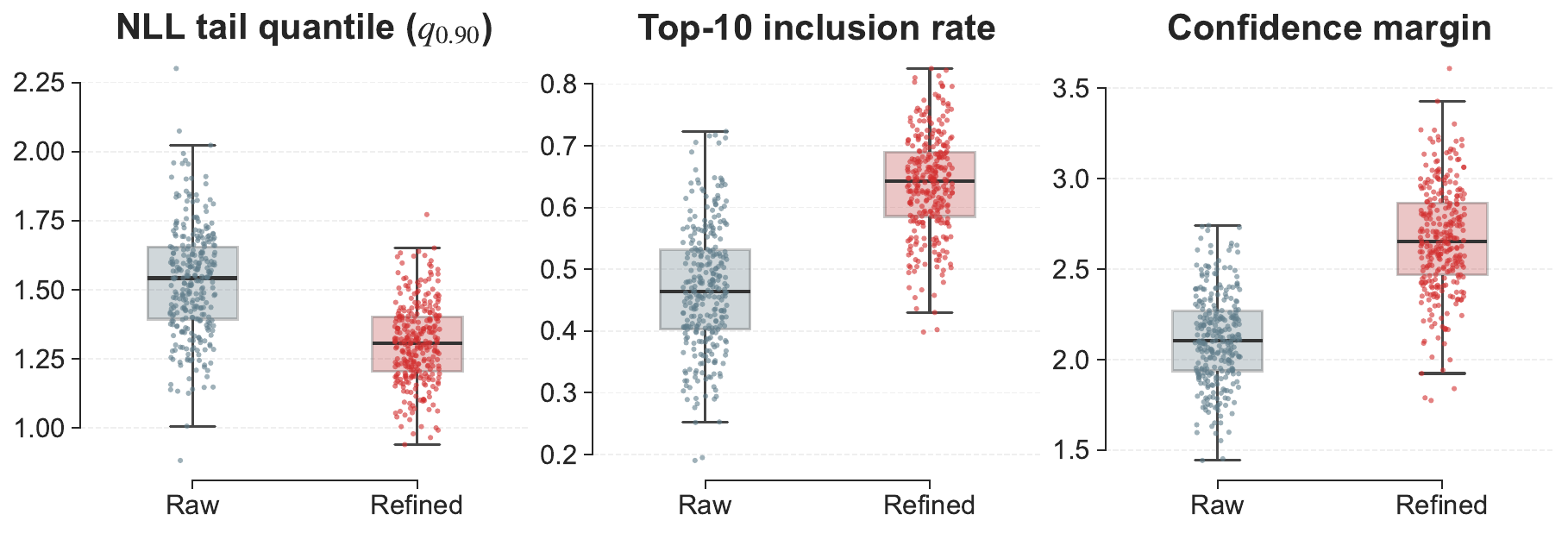}
  \caption{Feature diagnostics for teacher-forced logit.}
  \label{fig:feature_diagnostics}
  \vspace{-6mm}
\end{figure}

\section{Methodology}
\label{sec:method}

\begin{figure*}[t]
  \centering
  \includegraphics[width=\textwidth]{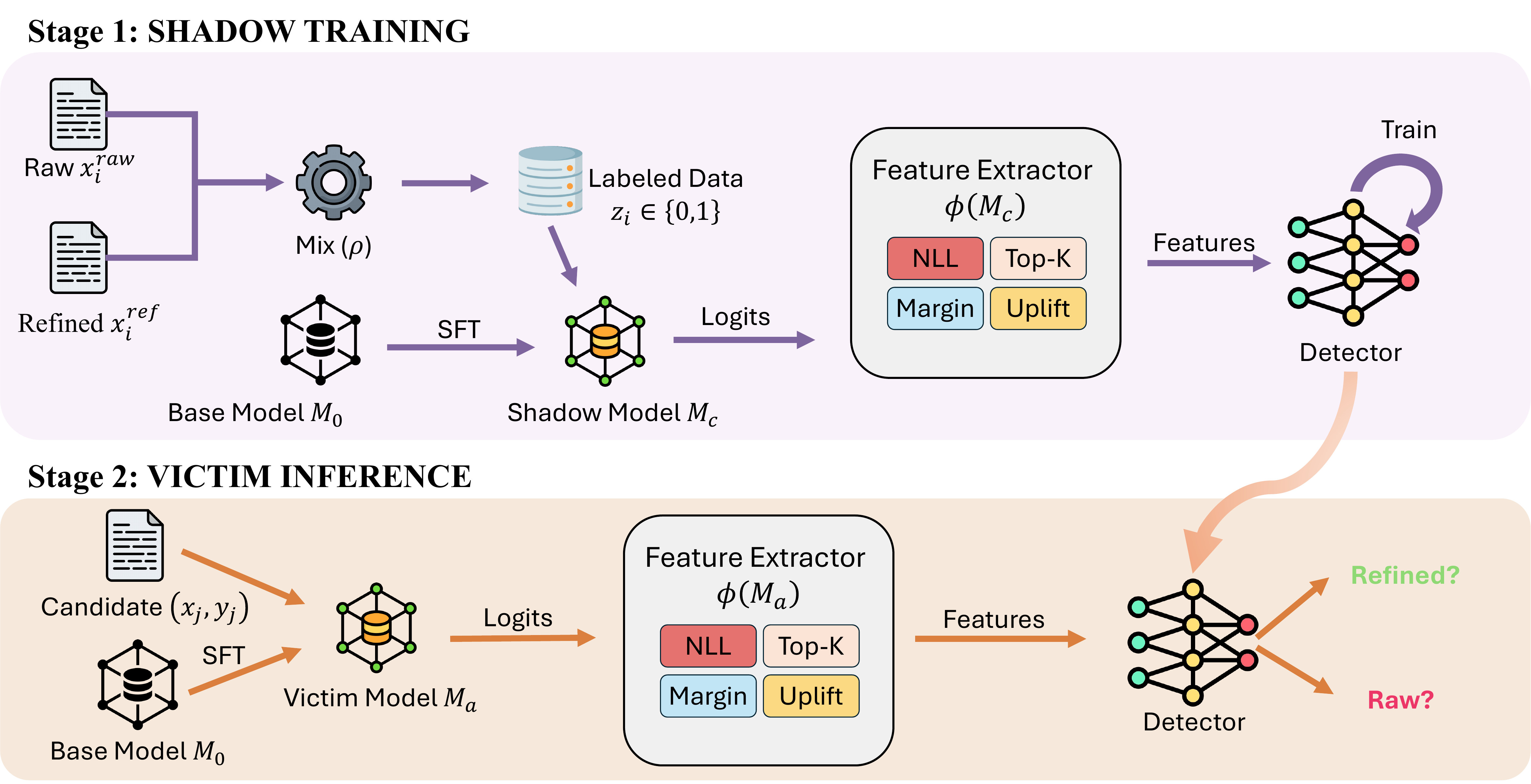}
  \caption{Overview of RePro. Stage~1 trains a supervised-contrastive encoder on logit-derived features from a shadow fine-tuned model. Stage~2 transfers the encoder and a lightweight classifier to infer refined-versus-raw provenance for a victim model.}
  \label{fig:overview}
  \vspace{-3mm}
\end{figure*}

Our goal is to infer whether a fine-tuned model $M_a$ was trained on a raw or an LLM-refined version of an instance. To operationalize this auditing objective, we develop an auditing attack and propose \textbf{RePro}, a framework that trains a supervised contrastive encoder on shadow fine-tuned models and transfers it to victims via a lightweight linear classifier. Figure~\ref{fig:overview} provides an overview of the full pipeline. In Stage~1, we construct a labeled shadow mixture of raw and refined instances, fine-tune a shadow model $M_c$ from the same base model $M_0$, and extract logit-derived feature vectors that summarize teacher-forced behavior through complementary signals such as NLL statistics, Top-$K$ ranking patterns, logit margins, and optional uplift features. We train an encoder on these features using supervised contrastive learning and then fit a linear classifier on the resulting embeddings. In Stage~2, we apply the same feature extraction and encoder to a victim model $M_a$ and use the transferred classifier to output a refined-versus-raw provenance score for each candidate instance.

\subsection{Teacher-Forced Logit Features}
Given a model $M$ and an instance $(x_i, y_i)$, we compute token-level log-probabilities under teacher forcing:
\begin{equation}
\ell^{(M)}_{i,t} = \log p_M(y_{i,t}\mid x_i, y_{i,<t}), \quad t=1,\dots,|y_i|.
\label{eq:logprob}
\end{equation}
Let $s^{(M)}_{i,t} \in \mathbb{R}^{|\mathcal{V}|}$ denote the pre-softmax logit vector at step $t$.
From $\{\ell^{(M)}_{i,t}\}_{t=1}^{|y_i|}$ and the corresponding logits, we derive a fixed-dimensional feature vector that summarizes (i) likelihood-based fit and tail hardness, and (ii) logit-based ranking behavior and local distribution sharpness.

\noindent
\textbf{Normalized negative log-likelihood (NLL).}
\begin{equation}
\mathrm{NLL}_M(i) = -\frac{1}{|y_i|}\sum_{t=1}^{|y_i|}\ell^{(M)}_{i,t}.
\label{eq:nll}
\end{equation}
To capture ``hard-token'' tails that are obscured by averaging, we additionally compute selected quantiles of tokenwise NLL values $\{-\ell^{(M)}_{i,t}\}$.

\noindent
\textbf{Top-$k$ inclusion.}
Let $\mathrm{TopK}_t^{(M)}$ denote the set of top-$k$ tokens under $M$ at step $t$ (ranked by logits in $s^{(M)}_{i,t}$). We define
\begin{equation}
\mathrm{TopK}_M(i) = \frac{1}{|y_i|}\sum_{t=1}^{|y_i|}\mathbb{I}\big[y_{i,t}\in \mathrm{TopK}_t^{(M)}\big],
\label{eq:topk}
\end{equation}
and use $k\in\{1,5,10\}$ in our experiments. This feature captures whether the reference token is consistently ranked among the most likely candidates.

\noindent
\textbf{Confidence margin.}
Let $s^{(M)}_{t,(1)}$ and $s^{(M)}_{t,(2)}$ be the largest and second-largest logit values in $s^{(M)}_{i,t}$, respectively. We compute the average margin
\begin{equation}
\mathrm{Gap}_M(i) = \frac{1}{|y_i|}\sum_{t=1}^{|y_i|}\left(s^{(M)}_{t,(1)} - s^{(M)}_{t,(2)}\right).
\label{eq:gap}
\end{equation}
This margin reflects the local sharpness of the next-token distribution and complements likelihood and ranking statistics.

\noindent
\textbf{Uplift.}
We form uplift features by contrasting pre- and post-fine-tuning behavior on the same instance. For any statistic $S(\cdot)\in\{\mathrm{NLL},\mathrm{TopK},\mathrm{Gap}\}$, we define
\begin{equation}
\Delta S(i) = S_{M_0}(i) - S_{M}(i),
\label{eq:uplift}
\end{equation}
where $M$ is the fine-tuned model of interest (victim $M_a$ or shadow $M_c$).
Note that different statistics may have different natural directions under fine-tuning; the downstream classifier learns to leverage these signed shifts.

We aggregate the above statistics into a feature vector
\begin{equation}
\phi(M; x_i, y_i) \in \mathbb{R}^d,
\label{eq:phi}
\end{equation}
which includes likelihood summaries, ranking signals, confidence geometry, and the corresponding uplift features from Eq.~\eqref{eq:uplift}. Prior to the next stage, we standardize each feature dimension using statistics computed on the shadow training split:
\begin{equation}
\tilde{\phi}_j = \frac{\phi_j - \mu_j}{\sigma_j + \epsilon},
\label{eq:standardize}
\end{equation}
with per-dimension mean $\mu_j$, standard deviation $\sigma_j$, and a small $\epsilon$ for numerical stability. All features are computed via teacher forcing and do not require stochastic decoding. In Figure~\ref{fig:feature_diagnostics}, we visualize these statistics on the same held-out instances and observe consistent distribution shifts between models fine-tuned on raw versus refined prompts.

\subsection{Shadow Training and Transfer}
To learn a transferable provenance classifier, we adopt a shadow fine-tuning setup. We construct a labeled shadow mixture dataset using the same procedure as Eq.~\eqref{eq:mixture}, yielding instances $\{(x_i, y_i, z_i)\}_{i\in\mathcal{I}_c}$ where $z_i$ indicates whether the prompt is refined, and fine-tune a shadow model $M_c$ from the same base model $M_0$. For each shadow instance $i$, we compute logit-derived features $\phi_i=\phi(M_c;x_i,y_i)$ and map them into an embedding space with an encoder $h_\psi$, i.e., $u_i=h_\psi(\phi_i)$.

We train $h_\psi$ using supervised contrastive learning: within each minibatch, embeddings with the same provenance label are pulled together while those with different labels are pushed apart,
\begin{equation}
\small
\min_{\psi}\ \sum_{i\in\mathcal{B}} \frac{-1}{|\mathcal{P}(i)|}\sum_{p\in \mathcal{P}(i)}
\log \frac{\exp(\mathrm{sim}(u_i,u_p)/\tau)}{\sum_{a\in \mathcal{B}\setminus\{i\}}\exp(\mathrm{sim}(u_i,u_a)/\tau)}
\label{eq:supcon}
\end{equation}
where $\mathcal{P}(i)=\{p\in\mathcal{B}\setminus\{i\}: z_p=z_i\}$ denotes positives, $\mathrm{sim}(\cdot,\cdot)$ is cosine similarity, and $\tau$ is a temperature hyperparameter. After contrastive training, we fit a lightweight linear classifier $g$ on top of the frozen embeddings $\{u_i\}$ using cross-entropy,
\begin{equation}
\min_{g}\ \sum_{i\in\mathcal{I}_c}\mathrm{CE}\!\left(g(u_i),\, z_i\right).
\label{eq:clf}
\end{equation}
At inference time, given a candidate instance $(x_j,y_j)$ for the victim model $M_a$, we compute $\phi(M_a;x_j,y_j)$, obtain $u_j=h_\psi(\phi(M_a;x_j,y_j))$, and output $g(u_j)$ as the refined-vs-raw provenance score.

\subsection{Complexity and Overhead}
\label{sec:method:complexity}
For each candidate instance $(x_i, y_i)$, RePro requires a single teacher-forced forward pass through the target model to obtain token-level log probabilities and top-$k$ logits along the reference sequence. Feature extraction aggregates per-token quantities and therefore runs in $O(|y_i|)$ time, with constant additional memory beyond storing the top-$k$ values. Applying the encoder $h_\psi$ and the linear classifier $g$ is negligible compared to the model forward pass. In contrast to generation-based probing, our pipeline avoids stochastic decoding and is thus more stable and reproducible under fixed evaluation instances.

\section{Experiments}
\label{sec:experiments}

\begin{table*}[t]
  \centering
    \caption{Main results across datasets, victims, and refiners. Each cell reports AUC / TPR@1\%FPR for RPI.}
    \vspace{-2mm}
  \small
  \setlength{\tabcolsep}{3.5pt}
  \renewcommand{\arraystretch}{1.15}
  \resizebox{\linewidth}{!}{
  \begin{tabular}{llcccccccc}
    \toprule
    \multirow{2}{*}{\textbf{Dataset}} &
    \multirow{2}{*}{\textbf{Victim}} &
    \multicolumn{4}{c}{\textbf{Refiner: GPT-4o}} &
    \multicolumn{4}{c}{\textbf{Refiner: Llama-3.3-70B-Instruct}} \\
    \cmidrule(lr){3-6} \cmidrule(lr){7-10}
    & & \textbf{$s_{\mathrm{NLL}}$} & \textbf{$s_{\Delta\mathrm{NLL}}$} & \textbf{$s_{\mathrm{pair}}$} & \textbf{RePro (Ours)}
      & \textbf{$s_{\mathrm{NLL}}$} & \textbf{$s_{\Delta\mathrm{NLL}}$} & \textbf{$s_{\mathrm{pair}}$} & \textbf{RePro (Ours)} \\
    \midrule

    \multirow{6}{*}{\textbf{GSM8K}}
      & Qwen2.5-1.5B-Instruct    & 0.53 / 0.07 & 0.58 / 0.10 & 0.57 / 0.09 & \textbf{0.66 / 0.16}
                                 & 0.52 / 0.06 & 0.57 / 0.09 & 0.56 / 0.08 & \textbf{0.64 / 0.14} \\
      & Qwen2.5-7B-Instruct      & 0.54 / 0.07 & 0.59 / 0.10 & 0.58 / 0.09 & \textbf{0.67 / 0.17}
                                 & 0.53 / 0.06 & 0.58 / 0.09 & 0.57 / 0.08 & \textbf{0.65 / 0.15} \\
      \cmidrule(lr){2-10}

      & Llama-3.1-8B-Instruct    & 0.55 / 0.08 & 0.60 / 0.11 & 0.59 / 0.10 & \textbf{0.69 / 0.19}
                                 & 0.54 / 0.07 & 0.59 / 0.10 & 0.58 / 0.09 & \textbf{0.67 / 0.17} \\
      & Llama-3.1-70B-Instruct   & 0.56 / 0.09 & 0.62 / 0.13 & 0.61 / 0.12 & \textbf{0.71 / 0.22}
                                 & 0.55 / 0.08 & 0.61 / 0.12 & 0.60 / 0.11 & \textbf{0.69 / 0.20} \\
      \cmidrule(lr){2-10}

      & Mistral-7B-Instruct-v0.3 & 0.54 / 0.07 & 0.59 / 0.10 & 0.58 / 0.09 & \textbf{0.67 / 0.17}
                                 & 0.53 / 0.06 & 0.58 / 0.09 & 0.57 / 0.08 & \textbf{0.65 / 0.15} \\
      & Mixtral-8x7B-Instruct-v0.1 & 0.55 / 0.08 & 0.61 / 0.12 & 0.60 / 0.11 & \textbf{0.70 / 0.20}
                                   & 0.54 / 0.07 & 0.60 / 0.11 & 0.59 / 0.10 & \textbf{0.68 / 0.18} \\
    \addlinespace[2pt]
    \midrule

    \multirow{6}{*}{\textbf{HumanEval}}
      & Qwen2.5-1.5B-Instruct    & 0.51 / 0.06 & 0.56 / 0.09 & 0.55 / 0.08 & \textbf{0.63 / 0.14}
                                 & 0.50 / 0.05 & 0.55 / 0.08 & 0.54 / 0.07 & \textbf{0.62 / 0.13} \\
      & Qwen2.5-7B-Instruct      & 0.52 / 0.06 & 0.57 / 0.09 & 0.56 / 0.08 & \textbf{0.65 / 0.15}
                                 & 0.51 / 0.05 & 0.56 / 0.08 & 0.55 / 0.07 & \textbf{0.63 / 0.13} \\
      \cmidrule(lr){2-10}

      & Llama-3.1-8B-Instruct    & 0.53 / 0.07 & 0.58 / 0.10 & 0.57 / 0.09 & \textbf{0.66 / 0.16}
                                 & 0.52 / 0.06 & 0.57 / 0.09 & 0.56 / 0.08 & \textbf{0.64 / 0.14} \\
      & Llama-3.1-70B-Instruct   & 0.54 / 0.08 & 0.60 / 0.12 & 0.59 / 0.11 & \textbf{0.68 / 0.18}
                                 & 0.53 / 0.07 & 0.59 / 0.11 & 0.58 / 0.10 & \textbf{0.66 / 0.16} \\
      \cmidrule(lr){2-10}

      & Mistral-7B-Instruct-v0.3 & 0.52 / 0.06 & 0.57 / 0.09 & 0.56 / 0.08 & \textbf{0.64 / 0.15}
                                 & 0.51 / 0.05 & 0.56 / 0.08 & 0.55 / 0.07 & \textbf{0.63 / 0.13} \\
      & Mixtral-8x7B-Instruct-v0.1 & 0.53 / 0.07 & 0.59 / 0.11 & 0.58 / 0.10 & \textbf{0.67 / 0.17}
                                   & 0.52 / 0.06 & 0.58 / 0.10 & 0.57 / 0.09 & \textbf{0.65 / 0.15} \\
    \bottomrule
  \end{tabular}
  }

  \label{tab:main_results_refiner}
\end{table*}

\begin{table*}[t!]
  \centering
    \caption{Cross-refiner generalization across victim families. Each cell reports AUC for RPI.}
\vspace{-2mm}
  \small
  \setlength{\tabcolsep}{6pt}
  \renewcommand{\arraystretch}{1.10}
  \begin{tabular}{llcc}
    \toprule
    \textbf{Victim} &
    \textbf{Train refiner (shadow)} $\rightarrow$ \textbf{Test refiner (victim)} &
    \textbf{GSM8K AUC} &
    \textbf{HumanEval AUC} \\
    \midrule

    \multirow{4}{*}{Qwen2.5-1.5B-Instruct}
      & GPT-4o $\rightarrow$ GPT-4o & 0.66 & 0.63 \\
      & GPT-4o $\rightarrow$ Llama-3.3-70B-Instruct & 0.65 & 0.65 \\
      & Llama-3.3-70B-Instruct $\rightarrow$ GPT-4o & 0.67 & 0.64 \\
      & Llama-3.3-70B-Instruct $\rightarrow$ Llama-3.3-70B-Instruct & 0.64 & 0.62 \\
    \midrule

    \multirow{4}{*}{Llama-3.1-8B-Instruct}
      & GPT-4o $\rightarrow$ GPT-4o & 0.69 & 0.66 \\
      & GPT-4o $\rightarrow$ Llama-3.3-70B-Instruct & 0.66 & 0.63 \\
      & Llama-3.3-70B-Instruct $\rightarrow$ GPT-4o & 0.67 & 0.64 \\
      & Llama-3.3-70B-Instruct $\rightarrow$ Llama-3.3-70B-Instruct & 0.67 & 0.64 \\
    \midrule

    \multirow{4}{*}{Mistral-7B-Instruct-v0.3}
      & GPT-4o $\rightarrow$ GPT-4o & 0.67 & 0.64 \\
      & GPT-4o $\rightarrow$ Llama-3.3-70B-Instruct & 0.66 & 0.65 \\
      & Llama-3.3-70B-Instruct $\rightarrow$ GPT-4o & 0.68 & 0.66 \\
      & Llama-3.3-70B-Instruct $\rightarrow$ Llama-3.3-70B-Instruct & 0.65 & 0.63 \\
    \bottomrule
  \end{tabular}

  \label{tab:cross_refiner}
\end{table*}

\begin{figure}[t]
  \centering
  \includegraphics[width=\linewidth]{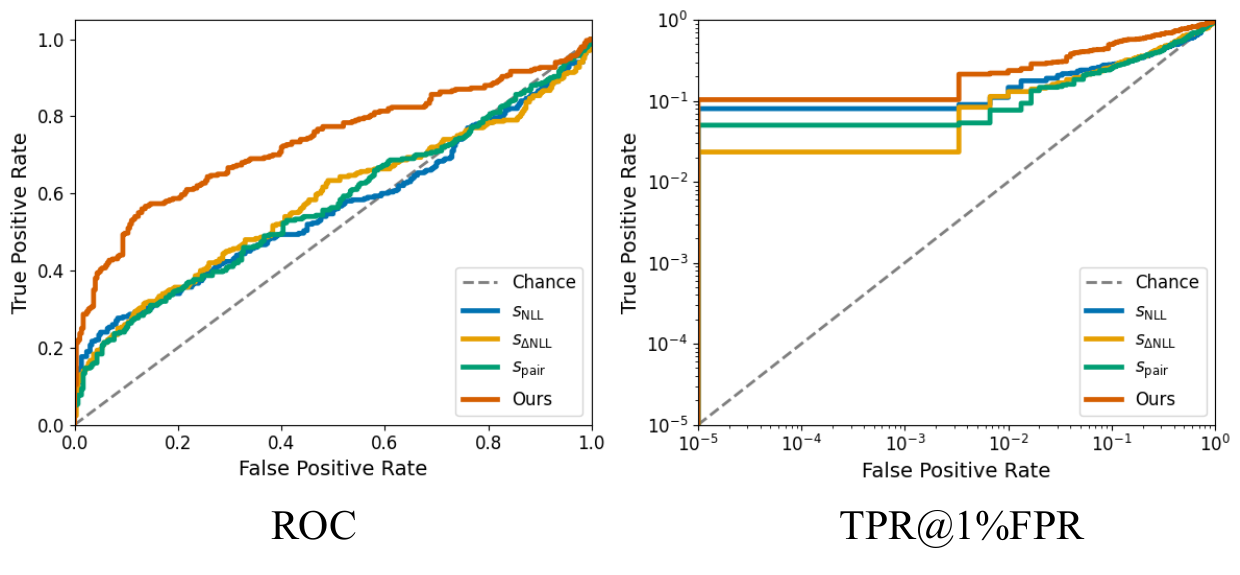}
  \caption{The ROC and TPR@1\% FPR curves.}
  \label{fig:roc}
  \vspace{-5mm}
\end{figure}

\begin{figure*}[t]
  \centering

  \begin{subfigure}[t]{0.32\linewidth}
    \centering
    \includegraphics[width=\linewidth]{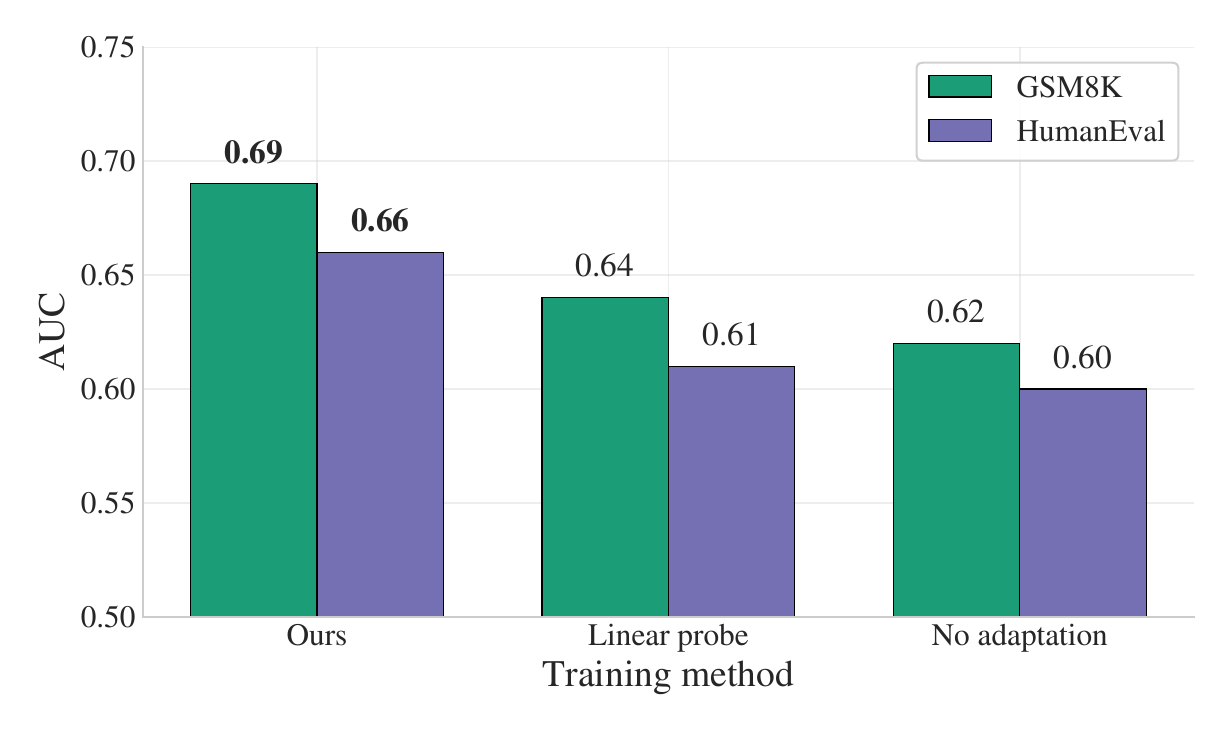}
    \caption{Training ablations}
    \label{fig:ablation_training}
  \end{subfigure}
  \hfill
  \begin{subfigure}[t]{0.32\linewidth}
    \centering
    \includegraphics[width=\linewidth]{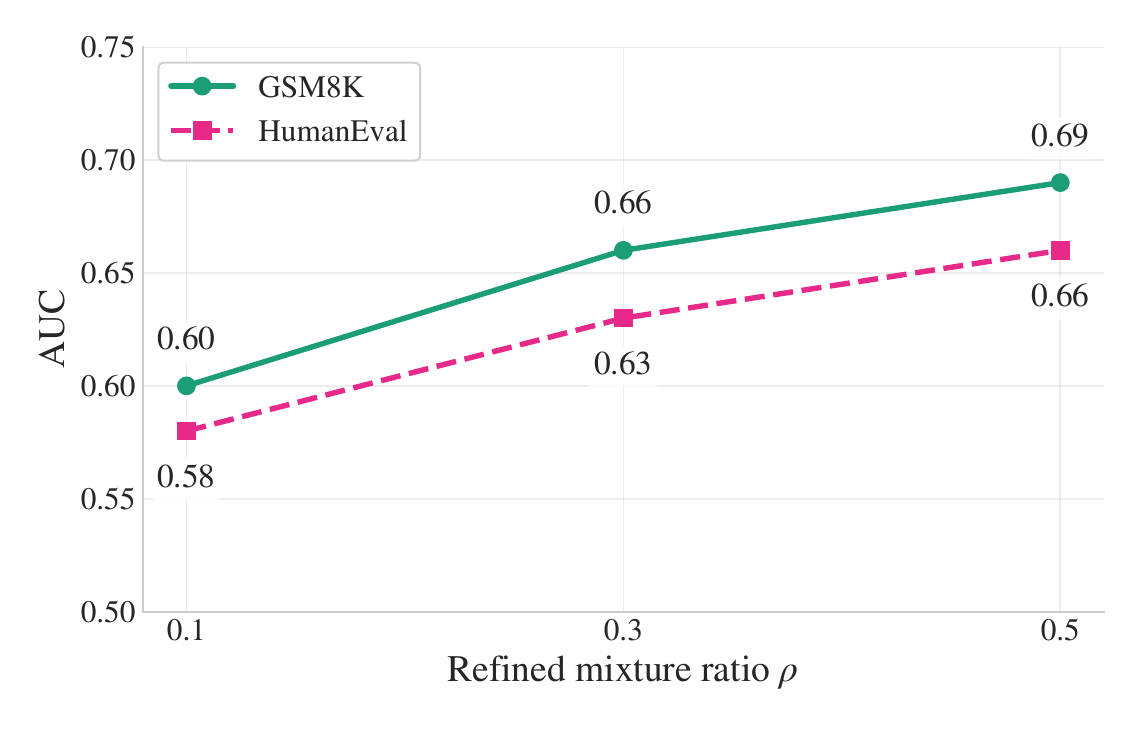}
    \caption{Sensitivity to refined fraction $\rho$}
    \label{fig:sensitivity_rho}
  \end{subfigure}
  \hfill
  \begin{subfigure}[t]{0.32\linewidth}
    \centering
    \includegraphics[width=\linewidth]{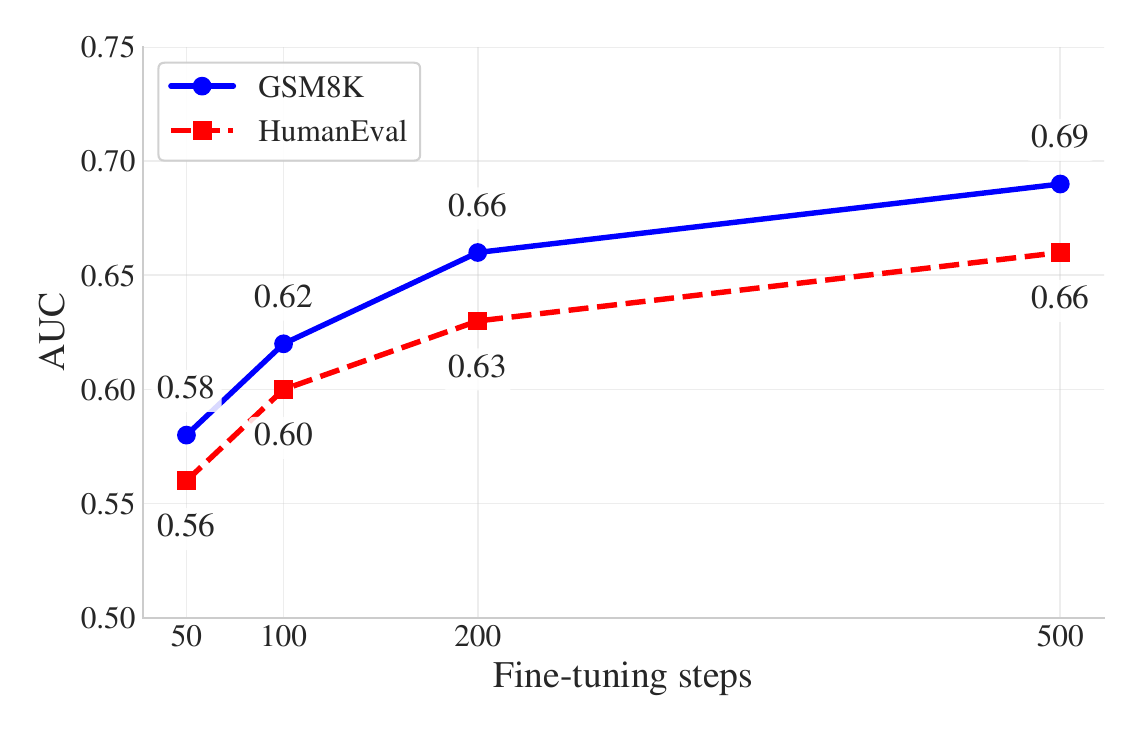}
    \caption{Sensitivity to fine-tuning steps}
    \label{fig:sensitivity_steps}
  \end{subfigure}

  \caption{Ablation and sensitivity analysis. 
  (a) Comparison of different attacker training strategies. 
  (b) AUC as a function of refined fraction $\rho$. 
  (c) AUC as a function of fine-tuning steps.}
  \label{fig:ablation_and_sensitivity}
\end{figure*}

We evaluate refinement provenance inference (RPI) on reasoning and code generation, testing whether a victim model fine-tuned on a mixture of raw and LLM-refined prompts exhibits detectable provenance traces in its teacher-forced token distributions, and whether the attacker generalizes across different refiners. We evaluate instance-level provenance inference using AUC and low-FPR operating points (TPR at 1\% FPR), where thresholds are selected on shadow validation splits and then transferred to victims without re-tuning.

\subsection{Experimental Setup}
\textbf{Datasets and refiners.}
We use GSM8K~\cite{cobbe2021gsm8k} for mathematical reasoning and HumanEval~\cite{chen2021evaluating} for code generation. For each semantic instance $i$, we take the dataset prompt as the raw prompt $x_i^{\mathrm{raw}}$ and the dataset-provided reference output as $y_i$, enabling teacher-forced logit extraction. We construct refined prompts $x_i^{\mathrm{ref}} = R(x_i^{\mathrm{raw}})$ using two refiners: a commercial LLM (GPT-4o~\cite{achiam2023gpt}) and an open-weight instruct model (Llama-3.3-70B-Instruct~\cite{dubey2024llama}). Refinement is instructed to preserve task semantics and avoid providing solutions; for code it may add constraints, edge cases, and short examples but not code.

\noindent
\textbf{Victims model and training configuration.}
Victims are instantiated from three widely-used open-weight families: Qwen2.5~\cite{qwen2.5}, Llama-3.1~\cite{dubey2024llama}, and Mistral~\cite{jiang2023mistral7b}. Starting from the corresponding base checkpoint $M_0$, we fine-tune the victim $M_a$ on a mixture with $\rho=0.5$, and train a shadow model $M_c$ on an instance-disjoint mixture constructed with the same $\rho$. Fine-tuning uses a fixed LoRA~\cite{hu2022lora} recipe across victim and shadow (rank $r{=}16$, $\alpha{=}32$, dropout $0.05$, learning rate $2\times10^{-4}$, 500 update steps, context length 768).

\noindent
\textbf{Attacker classifier.}
Our learned attacker is trained only on shadow data. Concretely, we map the logit feature vector $\phi(\cdot)$ to an embedding using a small MLP encoder $h_\psi$ (two fully-connected layers with ReLU, hidden size 256 and output size 128) followed by a 2-layer projection head (128$\rightarrow$64$\rightarrow$1). We train $h_\psi$ with a supervised contrastive objective on shadow instances, and then fit a linear classifier on top of the learned embeddings to predict refined vs.\ raw.

\noindent
\textbf{Learning-free baselines.}
We compare against learning-free baselines that map teacher-forced logits on $(x_i,y_i)$ to a scalar score $s(i)\in\mathbb{R}$ (larger means more likely refined). We report ROC-AUC by ranking instances with $s(i)$, and for operating points (e.g., $\mathrm{TPR}@\mathrm{FPR}{=}\alpha$) we threshold via $\hat z_i=\mathbb{I}[s(i)\ge \tau_\alpha]$ with $\tau_\alpha$ read from the empirical ROC curve. Concretely, we test: (i) victim-only likelihood $s_{\mathrm{NLL}}(i)=-\mathrm{NLL}_{M_a}(i)$; (ii) uplift likelihood $s_{\Delta\mathrm{NLL}}(i)=\mathrm{NLL}_{M_0}(i)-\mathrm{NLL}_{M_a}(i)$ when $M_0$ is available; (iii) pairwise preference $s_{\mathrm{pair}}(i)=\log p_{M_a}(y_i\mid x_i^{\mathrm{ref}})-\log p_{M_a}(y_i\mid x_i^{\mathrm{raw}})$.

\subsection{Matched-Refiner Evaluation}
\label{sec:exp:main-refiner}
We evaluate refinement provenance inference in a matched-refiner setting where the shadow attacker and the victim are constructed using the same refinement operator. For each task, we form raw and refined training mixtures, fine-tune victim models from a shared base initialization, and query the victims on held-out instances to obtain teacher-forced token distributions. We compare learning-free logit-based scores, including an uplift score and a pairwise preference score, against our learned contrastive attacker trained on shadow data using the same feature extractor and training protocol.

\noindent
\textbf{Result analysis.}
Table~\ref{tab:main_results_refiner} shows that the uplift and pairwise preference scores provide strong learning-free baselines, indicating that training on refined prompts leaves consistent traces in the victim’s token distributions beyond raw likelihood alone. Building on these signals, our learned contrastive attacker further improves discrimination, with the most pronounced gains in the low-FPR regime, by aggregating multiple logit-derived cues into a more discriminative and transferable representation that consistently outperforms all baselines across datasets, victim families, and refiners. Figure~\ref{fig:roc} shows the ROC and TPR@1\% FPR curves of the result with GPT-4o as refiner and Qwen2.5-1.5B-Instruct as victim.

\subsection{Cross-Refiner Transfer}
\label{sec:exp:cross-refiner}
To test whether provenance cues depend on the particular refinement operator, we conduct a cross-refiner transfer experiment that isolates refiner mismatch while varying the victim family. Specifically, we consider two refiners, GPT-4o and Llama-3.3-70B-Instruct, and for each victim family in Qwen2.5-1.5B-Instruct, Llama-3.1-8B-Instruct, Mistral-7B-Instruct-v0.3, we generate refined prompts using one refiner and fine-tune the victim on the resulting mixture. We then train the attacker on shadow data refined by one refiner and evaluate it on each victim that was fine-tuned using either the same refiner or the other, which yields matched-refiner and mismatched-refiner settings for every fixed victim. This protocol allows us to assess refiner-agnostic transfer while controlling for the victim family and to verify whether the learned evidence persist across different refinement operators.

\noindent
\textbf{Result analysis.}
Table~\ref{tab:cross_refiner} shows that performance remains strong in the mismatched cases with only a moderate degradation relative to the matched setting, suggesting that the attacker leverages refiner-agnostic cues that reflect distribution-level preference shifts induced by refined-prompt training rather than artifacts specific to any single refiner.

\subsection{Ablation Study}
\label{sec:exp:ablation}

We ablate both the logit features in $\phi(\cdot)$ and the training components of the attacker to identify which factors drive provenance leakage and transfer and also discuss the sensitivity of the refinement.

\noindent
\textbf{Feature ablations.}
Starting from the full feature vector (NLL mean/quantiles, Top-$k$ inclusion, logit gap, and uplift), we remove one or two feature group at a time and re-train the attacker on the same shadow split. Table~\ref{tab:ablation_features} reports the resulting AUC and the absolute drop relative to the full model. Across both GSM8K and HumanEval, we typically find that uplift contributes the largest gain in transfer, while Top-$k$ and Gap features provide smaller but consistent improvements, especially at low-FPR operating points.

\begin{table}[t]
  \centering
    \caption{Feature ablation. We report AUC and the absolute drop relative to the full feature set.}
    \vspace{-2mm}
  \small
  \setlength{\tabcolsep}{5pt}
  \renewcommand{\arraystretch}{1.12}
  \begin{tabular}{lcc}
    \toprule
    \textbf{Variant} & \textbf{GSM8K} & \textbf{HumanEval} \\
    \midrule
    w/o uplift & 0.65 \,(-0.04) & 0.63 \,(-0.03) \\
    w/o NLL tails & 0.68 \,(-0.01) & 0.65 \,(-0.01) \\
    w/o ranking & 0.67 \,(-0.02) & 0.64 \,(-0.02) \\
    w/o margin & 0.66 \,(-0.03) & 0.65 \,(-0.01) \\

    \midrule
    w/o uplift + NLL tails & 0.60 \,(-0.09) & 0.58 \,(-0.08) \\
    w/o uplift + ranking & 0.57 \,(-0.12) & 0.55 \,(-0.11) \\
    w/o uplift + margin & 0.58 \,(-0.11) & 0.59 \,(-0.07) \\
    w/o NLL tails + ranking & 0.56 \,(-0.13) & 0.57 \,(-0.09) \\
    w/o NLL tails + margin & 0.55 \,(-0.14) & 0.58 \,(-0.08) \\
    w/o ranking + margin & 0.60 \,(-0.09) & 0.60 \,(-0.07) \\

    \midrule
    \textbf{Ours} & \textbf{0.69} & \textbf{0.66} \\
    \bottomrule
  \end{tabular}

  \label{tab:ablation_features}
\end{table}

\noindent
\textbf{Attacker training ablations.}
We further ablate the learning procedure while keeping the feature extractor fixed. Specifically, we compare our supervised-contrastive training to: (i) linear probe only (train a linear classifier directly on the raw feature vector $\phi$ without representation learning), and (ii) no adaptation (use a shadow model without fine-tuning, i.e., replace $M_c$ with the base model $M_0$). Figure~\ref{fig:ablation_training} summarizes performance, showing that contrastive training improves robustness by shaping an embedding where refined-vs-raw separation transfers better across victims and refiners.

\noindent
\textbf{Sensitivity to refinement strength.}
Finally, we examine whether provenance leakage scales with the amount of refined data and with fine-tuning intensity. We sweep the refined mixture ratio $\rho\in\{0.1,0.3,0.5\}$ and the fine-tuning budget (number of update steps). Figure~\ref{fig:sensitivity_rho} and figure~\ref{fig:sensitivity_steps} plots AUC as a function of $\rho$ and training steps, respectively. As $\rho$ increases or fine-tuning becomes stronger, the refined distribution contributes a larger fraction of gradient updates, typically amplifying the preference shift and increasing detectability.

\begin{figure}[t]
  \centering
  \includegraphics[width=\linewidth]{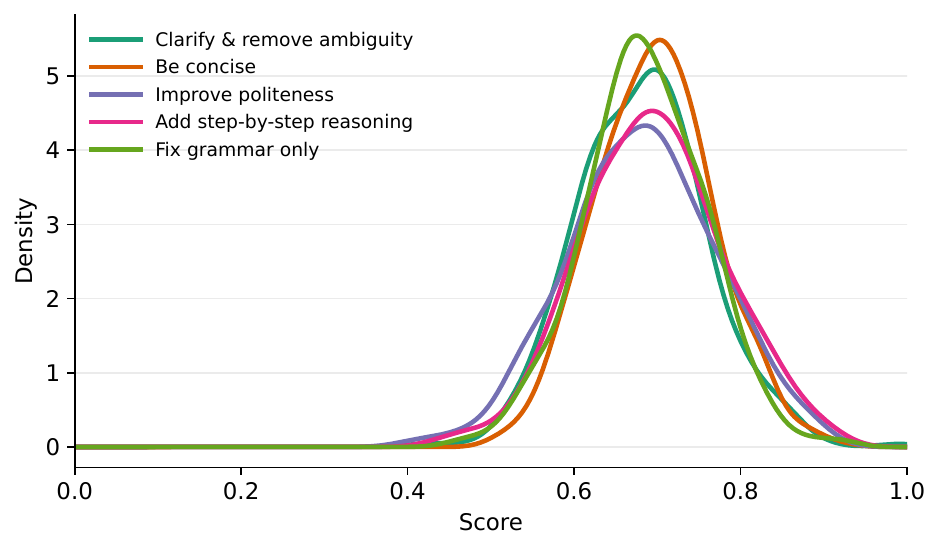}
  \caption{Score distributions under different refinement instruction templates.}
  \label{fig:refine_instr_dist}
  \vspace{-5mm}
\end{figure}

\noindent
\textbf{Analysis of refinement template.}
We further analyze the distribution of the attacker’s classification score $g(x)$ for predicting whether $x$ is refined or raw under different refinement instruction templates.
For each template $c$, we compute $g(x)$ for all evaluation instances and estimate $\hat{p}(g\mid c)$ via KDE.
As shown in Figure~\ref{fig:refine_instr_dist}, the score distributions are highly consistent across instruction variants, indicating that our decision signal is not tied to a specific rewriting style and remains stable under instruction-level variations.

\section{Conclusion}
\label{sec:conclusion}
We propose Refinement Provenance Inference, which asks whether a fine-tuned language model was trained on raw prompts or prompts rewritten by an external refiner LLM. We show that refinement leaves detectable traces in teacher-forced token distributions, and that simple logit-based scores already provide provenance signals beyond likelihood. Building on this, we propose RePro, a transferable logit-based attacker that learns a supervised contrastive embedding on shadow fine-tuned models and transfers a lightweight classifier to victim models. Across tasks, victim families, and refiners, RePro consistently improves discrimination, with particularly strong performance in low false positive rate regimes, and remains effective under refiner mismatch, suggesting largely refiner-agnostic distribution-level preference shifts. Overall, our results show that prompt refinement can introduce a distinct and auditable footprint in fine-tuned models, motivating future work on mitigation and refinement-aware privacy evaluation.

\section*{Limitations}
\label{sec:limitations}
Our study focuses on refinement provenance inference under a teacher-forcing interface and therefore inherits several limitations. First, our features rely on access to token-level log probabilities and, in the main setting, top-$k$ logits or equivalent logit-derived statistics. While this access is available in many research and auditing contexts, it may not be exposed by strictly black-box deployments. Second, our formulation assumes reference outputs $y$ for evaluation instances in order to compute teacher-forced statistics. This matches supervised benchmarks such as GSM8K and HumanEval, but it may be restrictive in fully open-ended settings where gold references are unavailable or ambiguous. Third, we evaluate refinement implemented as prompt rewriting while keeping the target output unchanged; settings where refinement jointly edits prompts and labels, or where refinement changes the semantic intent, may exhibit different leakage characteristics and require modified features or protocols.

\bibliography{custom}

\appendix
\newpage
\section{Refinement Instructions and Templates}

\subsection{GSM8K Refinement Prompt}
\begin{tcolorbox}[
    colback=gray!3,
    colframe=black!70,
    boxrule=0.55pt,
    arc=0.8mm,
    left=2mm,right=2mm,top=1.3mm,bottom=1.3mm
]
\ttfamily\small
\textbf{System:} Rewrite the prompt for instruction tuning: improve clarity/structure, preserve semantics.\\
\textbf{Rules:} Do not solve or hint. No reasoning/derivations/formulas/answers. Preserve all quantities/conditions and the required output. Output only the rewritten prompt.\\[0.3em]
\textbf{User:} Rewrite this GSM8K word problem into a clear instruction. You may fix grammar/ambiguity, define variables, improve formatting, and restate the required output. Do not include any solution steps or computed results.\\
\textbf{RAW:} \verb|<<< {X_RAW} >>>|
\end{tcolorbox}

\subsection{HumanEval Refinement Prompt}
\begin{tcolorbox}[
    colback=gray!3,
    colframe=black!70,
    boxrule=0.55pt,
    arc=0.8mm,
    left=2mm,right=2mm,top=1.3mm,bottom=1.3mm
]
\ttfamily\small
\textbf{System:} Rewrite the code-task prompt to improve clarity/completeness, preserve semantics.\\
\textbf{Rules:} Do not implement. No code or pseudocode. If the raw prompt contains code (e.g., signature/stub), keep it exactly; only edit surrounding natural-language text. You may add constraints, edge cases, and brief plain-text I/O examples. Output only the rewritten prompt.\\[0.3em]
\textbf{User:} Rewrite this HumanEval task into a clearer specification: intent, inputs/outputs, constraints, corner cases, and brief plain-text examples if helpful. Do not provide implementation details or any code/pseudocode.\\
\textbf{RAW:} \verb|<<< {X_RAW} >>>|
\end{tcolorbox}

\section{Data Construction and Disjointness Protocol}

We construct victim and shadow fine-tuning corpora using the same raw/refined mixture protocol, while enforcing strict instance-level disjointness between victim and shadow training data. This ensures the attacker learns provenance cues that transfer beyond memorizing specific prompts.

\begin{table}[t!]
  \centering
  \caption{Data construction protocol for victim and shadow corpora (instance-disjoint).}
  \vspace{-2mm}
  \small
  \setlength{\tabcolsep}{6pt}
  \renewcommand{\arraystretch}{1.12}
  \begin{tabular}{p{0.36\linewidth} p{0.57\linewidth}}
    \toprule
    \textbf{Item} & \textbf{Protocol} \\
    \midrule
    Disjointness unit & Dataset instance (problem / function) \\
    Victim pool & $\mathcal{D}_v$ (no overlap with $\mathcal{D}_s$) \\
    Shadow pool & $\mathcal{D}_s$ (no overlap with $\mathcal{D}_v$) \\
    Mixture indicator & $z_i \sim \mathrm{Bernoulli}(\rho)$ per instance \\
    Mixture fixing & Sample $z_i$ once; keep fixed across training \\
    Prompt form & $x_i = x_i^{\mathrm{raw}}$ if $z_i=0$; else $x_i^{\mathrm{ref}}$ \\
    Refinement caching & Single rewrite per $x_i^{\mathrm{raw}}$; cached thereafter \\
    Label handling & Keep reference output $y_i$ unchanged \\
    Validation split & Held-out subset from each pool \\
    Evaluation split & Held-out set disjoint from all fine-tuning instances \\
    Length handling & Apply the same tokenization/truncation rules to all sets \\
    Randomness control & Fixed random seed for splits and $z_i$ sampling \\
    \bottomrule
  \end{tabular}
  \label{tab:data_disjointness}
\end{table}

\section{Victim and Shadow Fine-tuning Details}

Victim models are fine-tuned from a base checkpoint on a mixture of raw and refined prompts.
Shadow models use the same fine-tuning recipe but are trained on an instance-disjoint mixture
constructed with the same protocol, enabling transferable attacker training. The specific setting can be seen from Table~\ref{tab:ft_recipe}.

\begin{table}[t!]
  \centering
  \caption{Shared fine-tuning configuration for victim and shadow models.}
  \vspace{-2mm}
  \small
  \setlength{\tabcolsep}{6pt}
  \renewcommand{\arraystretch}{1.12}
  \begin{tabular}{p{0.33\linewidth} p{0.58\linewidth}}
    \toprule
    \textbf{Training component} & \textbf{Setting (shared by victim and shadow)} \\
    \midrule
    Fine-tuning objective & Supervised fine-tuning (SFT) on $(x, y)$ pairs \\
    Parameter-efficient tuning & LoRA \\
    LoRA rank $r$ & 16 \\
    LoRA scaling $\alpha$ & 32 \\
    LoRA dropout & 0.05 \\
    Learning rate & $2 \times 10^{-4}$ \\
    Training steps & 500 updates \\
    Context length & 768 tokens \\
    Mixture rate (main) & $\rho = 0.5$ \\
    \bottomrule
  \end{tabular}
  \label{tab:ft_recipe}
\end{table}

\section{Future Work}
\label{app:future_work}
Future work can extend refinement provenance inference in several directions. One is to audit richer curation pipelines beyond single-pass prompt rewriting, such as multi-turn refinement or joint prompt-and-response transformations, to understand which cues remain stable under more complex operators. Another is to relax the reliance on teacher-forced statistics with a known reference output, enabling auditing with weaker interfaces such as sampled generations or score-only APIs. Finally, it is important to study adaptive obfuscation and mitigation, including mixing refiners or style randomization to reduce distinguishability, and to evaluate the resulting privacy.



\end{document}